\documentclass[letterpaper, 10 pt, conference]{packages/ieeeconf}

\IEEEoverridecommandlockouts                              

\usepackage{times}
\usepackage{epsfig}
\usepackage{graphicx}
\usepackage{amsmath}
\usepackage{amssymb}
\usepackage{subfigure} 
\usepackage[ruled]{algorithm}
\usepackage{algpseudocode}
\usepackage{ctable}
\usepackage{url}
\usepackage[bookmarks=true]{hyperref}
\usepackage[export]{adjustbox}
\usepackage{import}
\usepackage{pgf}
\usepackage{setspace}

\makeatletter
\let\NAT@parse\undefined
\makeatother
\usepackage[numbers]{natbib}

\usepackage{packages/sparo_acronyms}
\usepackage{packages/sparo_math}
\usepackage{packages/sparo_SIunits}
\usepackage{packages/sparo_misc}
\usepackage{multirow}

\usepackage{soul,color}
\usepackage{verbatim} 

\usepackage{array}
\newcolumntype{L}[1]{>{\raggedright\let\newline\\\arraybackslash\hspace{0pt}}m{#1}}
\newcolumntype{C}[1]{>{\centering\let\newline\\\arraybackslash\hspace{0pt}}m{#1}}
\newcolumntype{R}[1]{>{\raggedleft\let\newline\\\arraybackslash\hspace{0pt}}m{#1}}

\usepackage{xcolor}

\usepackage[symbol]{footmisc}

\title{\LARGE \bf
    Mesh-based Photorealistic and Real-time 3D Mapping \\ for Robust Visual Perception of Autonomous Underwater Vehicle
}

\author{Jungwoo Lee$^{1}$ and Younggun Cho$^{1*}$
\thanks{This work was supported by the National Research Foundation of Korea(NRF) grant funded by the Korea government(MSIP) (No.2022R1A4A3029480 and RS-2023-00302589) and Institute of Information \& communications Technology Planning \& Evaluation (IITP) grant funded by the Korea government(MSIT) (No.2022-0-00448).}
\thanks{$^{1}$Jungwoo Lee and $^{1*}$Younggun Cho is with the Dept. Electrical and Computer Engineering, Inha University, Incheon, South Korea
        {\tt\small pihsdneirf@inha.edu}, {\tt\small yg.cho@inha.ac.kr}}%
}

\usepackage{fancyhdr}
\fancypagestyle{withfooter}{
  
  \fancyfoot[C]{\footnotesize Accepted to the IEEE ICRA Workshop on Field Robotics 2024}
}

\begin{document}

\maketitle
\thispagestyle{withfooter}
\pagestyle{withfooter}

\begin{abstract} 
This paper proposes a photorealistic real-time dense 3D mapping system that utilizes a learning-based image enhancement method and mesh-based map representation. Due to the characteristics of the underwater environment, where problems such as hazing and low contrast occur, it is hard to apply conventional \ac{SLAM} methods. Furthermore, for sensitive tasks like inspecting cracks, photorealistic mapping is very important. However, the behavior of \ac{AUV} is computationally constrained. In this paper, we utilize a neural network-based image enhancement method to improve pose estimation and mapping quality and apply a sliding window-based mesh expansion method to enable lightweight, fast, and photorealistic mapping. To validate our results, we utilize real-world and indoor synthetic datasets. We performed qualitative validation with the real-world dataset and quantitative validation by modeling images from the indoor synthetic dataset as underwater scenes.    
\end{abstract} 
\section{Introduction}

With the recent evolution of marine robotics, there is growing interest in SLAM research that utilizes robots to perform various tasks in underwater environments. This research is being utilized in various fields, such as underwater construction and infrastructure management, ocean environment analysis, and underwater mapping. In particular, the utilization of \ac{AUV} is emphasized for hazardous underwater terrains that are dangerous for human operators and structures that require constant monitoring for safety. In this paper, we propose a robust localization and photorealistic dense mapping system to perform sensitive tasks such as inspecting cracks in underwater infrastructure.

Visual information is essential to creating realistic maps easily interpreted by humans. However, utilizing image data in an underwater environment is challenging. In underwater environments, image quality is degraded by light scattering and absorption, which differs from atmospheric conditions. Image quality degradation, such as color distortion and contrast reduction, makes it difficult for humans to interpret the image information and affects the \ac{SLAM} performance of underwater robots.

\begin{figure}[t] 
    \centering
    \includegraphics[width=\columnwidth]{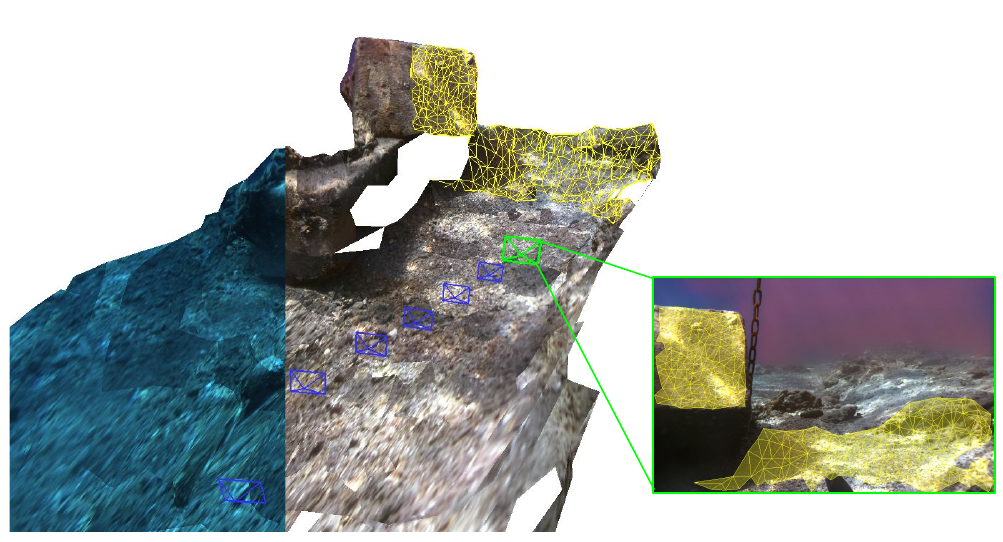}
    \caption{Illustration of the proposed method. We utilize a learning-based image enhancement and apply mesh-based mapping to generate a photorealistic map.} 
    \label{fig:thumnail}
\end{figure}

To solve this problem, various methods have been proposed for color correction of underwater images \cite{cho2016onl}. Recently, with the development of artificial intelligence, many neural network-based image enhancement methods have been presented \cite{jointid}. As in \figref{fig:thumnail}, the system proposed in this paper employs a learning-based image enhancement method to enhance the localization performance of underwater robots and simultaneously generate easy-to-interpret maps for humans.

\begin{table}
\caption{Comparison of map representation.}
\begin{center}
\begin{tabular}{l||c|c|c}
    \hline
    Representation    & Realistic & Speed & Difficulty \\
    \hline\hline
    Point Cloud  & Middle        & Middle        & \textbf{Low} \\
    \hline
    Voxel       & Low           & \textbf{High} & Middle \\
    \hline
    Mesh        & \textbf{High} & \textbf{High} & High  \\
    \hline
\end{tabular}
\end{center}
\label{tab:compmaprep}
\end{table}

\begin{figure*}[t!] 
    \centering
    \includegraphics[width=0.9\textwidth]{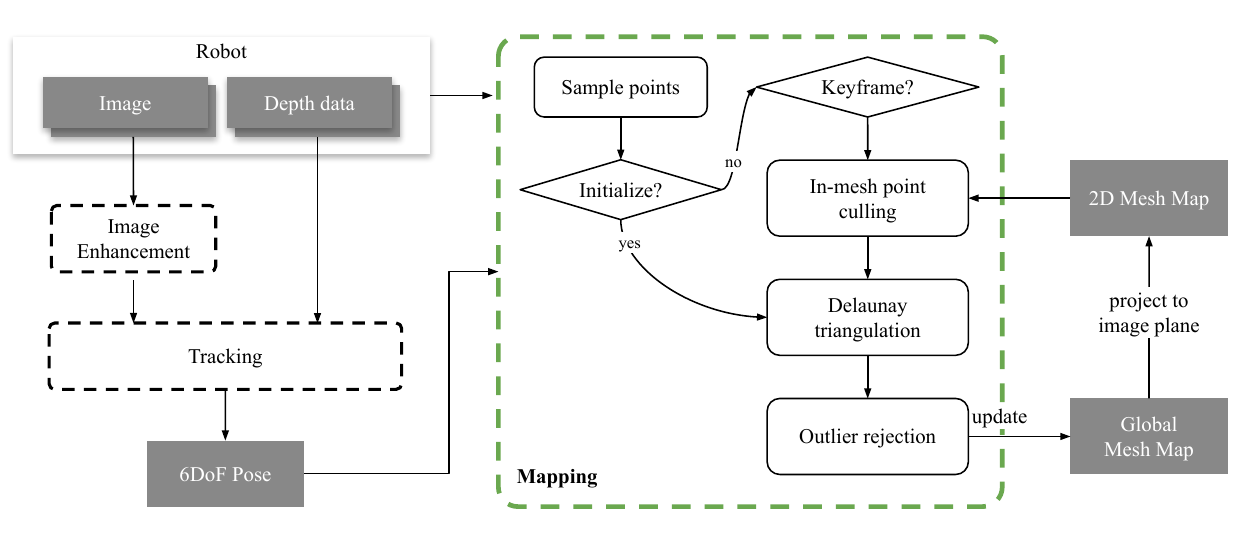}
    \caption{The overall pipeline of the proposed system.}
    \label{fig:pipeline}
\end{figure*}

To create realistic and easy-to-interpret maps, dense mapping is essential, and map representations such as point cloud \cite{kinectfusion}, voxel \cite{voxblox}, and mesh \cite{kimeravio} are commonly used. Point clouds are easy to handle and have realistic information about specific points, but they have limitations in that they cannot express continuous positional information. On the other hand, voxels allow us to control the size of specific areas according to the predefined size of the block. Still, it is necessary to sacrifice realistic representation for real-time operation. Another approach is a mesh-based map representation. Mesh is an unstructured representation, which makes it difficult to handle data. However, it has the advantage of using a small amount of data while still providing a realistic representation through texture mapping. See \tabref{tab:compmaprep} for a comparison of map representation.

This paper proposes a system for robust localization and realistic mapping in underwater environments with the following contributions.
\begin{itemize}
\item Improving localization performance of robots and map representation by applying neural network-based underwater image enhancement methods.
\item Real-time pixel-by-pixel map generation through the employing mesh-based map representation.
\item Qualitative validation of the proposed method using real-world datasets and quantitative validation using indoor synthetic datasets.
\end{itemize}
\section{Method}

\subsection{System Overview}

\begin{figure}[t] 
    \centering
    \subfigure[Simplified illustration of the Joint-ID's network \cite{jointid}.]{
        \includegraphics[width=\columnwidth]{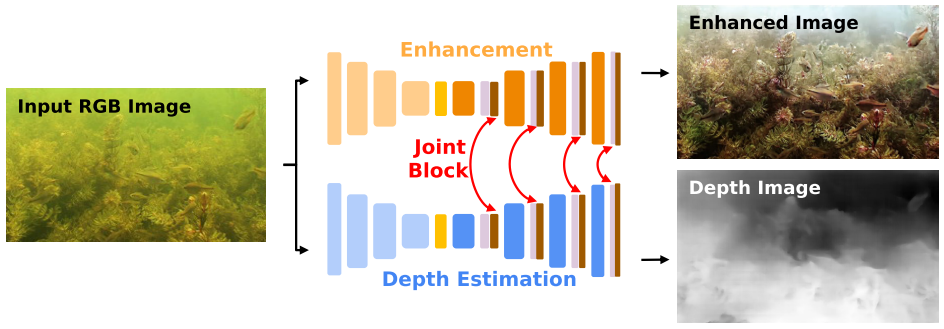}
        \label{fig:jointid:pipeline}
    }
    \subfigure[Hazed image.]{
        \includegraphics[width=0.45\columnwidth]{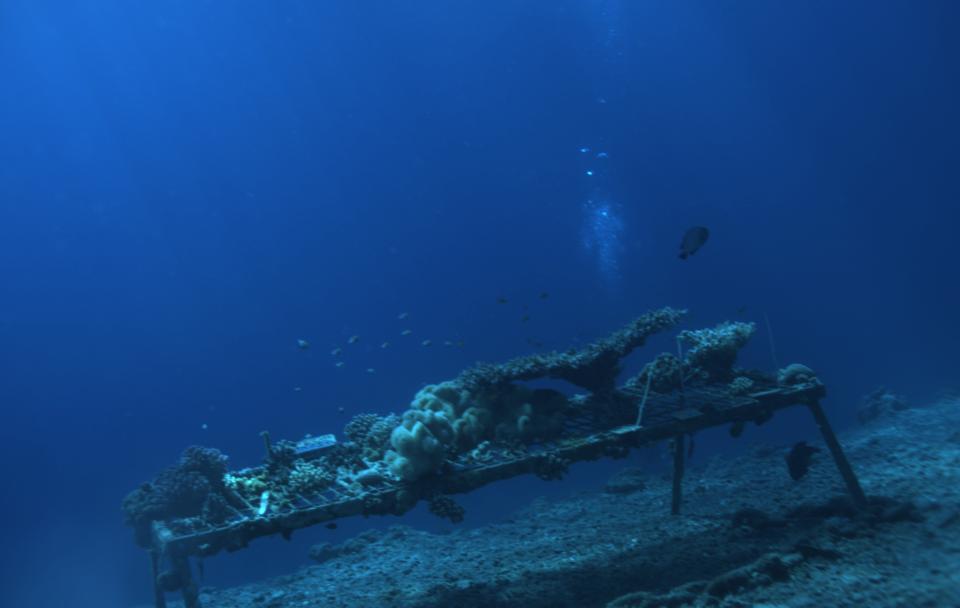}
        \label{fig:jointid:hazed}
    }
    \subfigure[Enhanced image.]{
        \includegraphics[width=0.45\columnwidth]{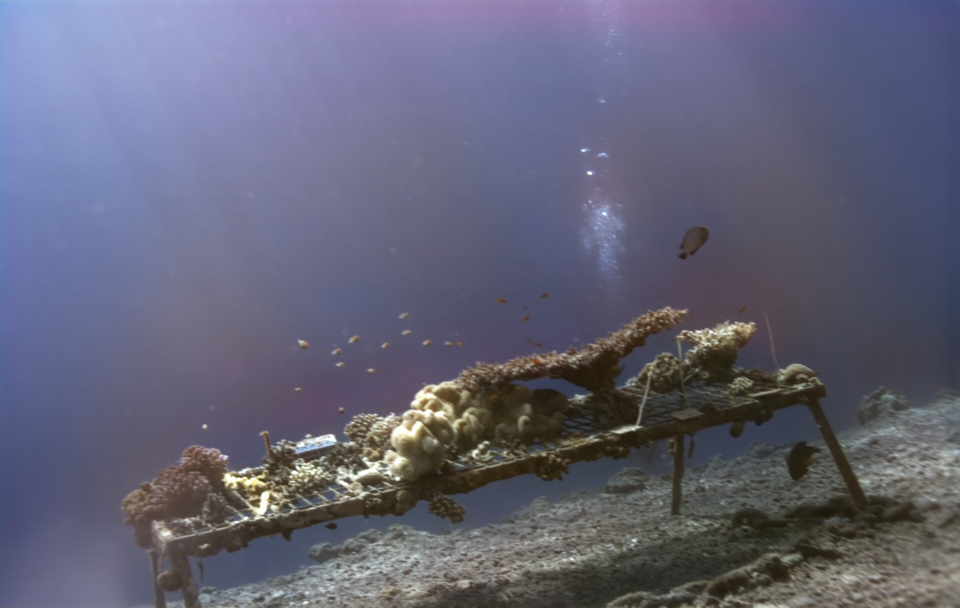}
        \label{fig:jointid:enhanced}
    }
    
    \caption{Illustration of Joint-ID and image enhancement result.}
    \label{fig:jointid}
\end{figure}

\figref{fig:pipeline} shows the overall system pipeline of the proposed system. First, it performs underwater image enhancement. In this paper, we utilize Joint-ID \cite{jointid}, which is based on the transformer model, and it shows \ac{SOTA} performance for underwater image enhancement. The image enhancement results can be seen in \figref{fig:jointid:hazed} and \figref{fig:jointid:enhanced}. Subsequently, the enhanced image and depth information are utilized to estimate the robot's pose. The localization was performed based on the ORB-SLAM2 \cite{orbslam2}. Finally, the enhanced image, depth, and pose information of the robots are used to generate a dense 3D mesh map. The details of dense 3D mesh mapping is described in \secref{sec:densemapping}.

\subsection{Neural Network-Based Underwater Image Enhancement}

In this paper, we utilized Joint-ID \cite{jointid} for underwater image enhancement. Joint-ID is a transformer-based neural network model that performs image enhancement and depth estimation simultaneously from a single image, in contrast to other existing deep learning approaches that perform either image enhancement or depth estimation separately. And this model shows much more powerful performance than existing methods. The structure of the neural network can be seen in \figref{fig:jointid:pipeline}.

\subsection{Dense 3D Mesh Mapping}
\label{sec:densemapping}

\subsubsection{3D Mesh Generation}

\begin{figure}[t] 
    \centering
    \includegraphics[width=\columnwidth,clip,trim={0 0 0 1.5cm}]{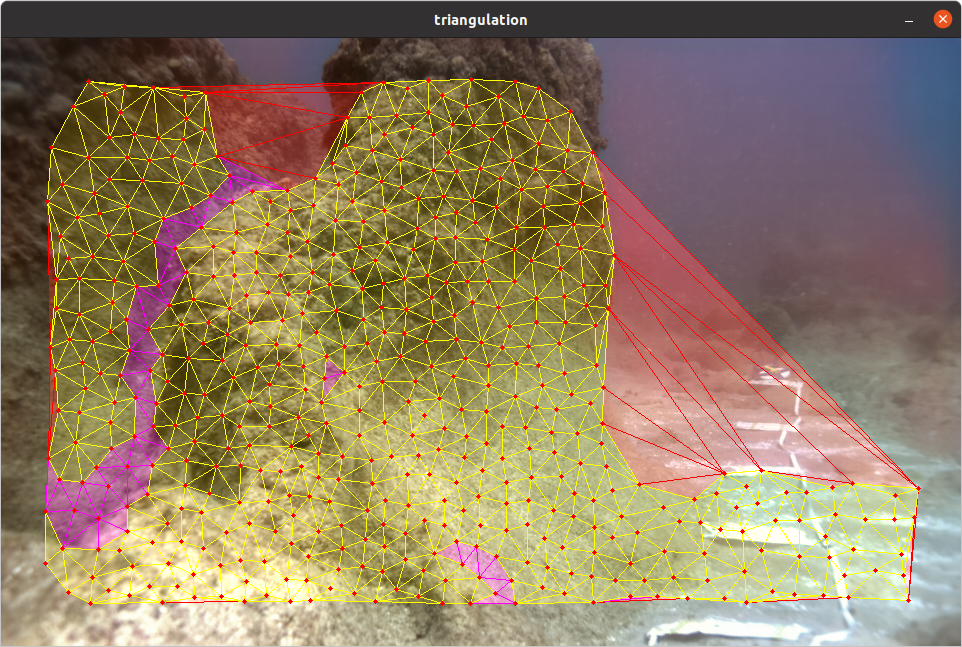}
    \caption{Point sampling and triangulation: The red points are extracted feature points. We performed Delaunay triangulation on the extracted points. The red triangle mesh is the one that is removed from the image plane. The purple triangle mesh is removed after the 3D projection. The yellow triangle mesh is the correct one on the map.}
    \label{fig:triangulation}
\end{figure}

\begin{figure}[t] 
    \centering
    \subfigure[Before outlier rejection.]{
        \includegraphics[width=0.45\columnwidth,frame]{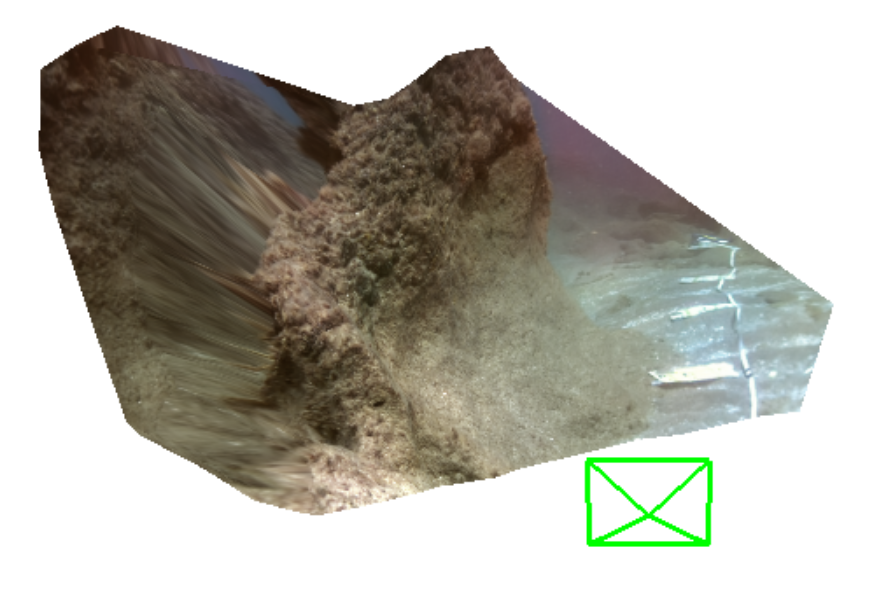}
        \label{fig:outlier:before}
    }
    \subfigure[After outlier rejection.]{
        \includegraphics[width=0.45\columnwidth,frame]{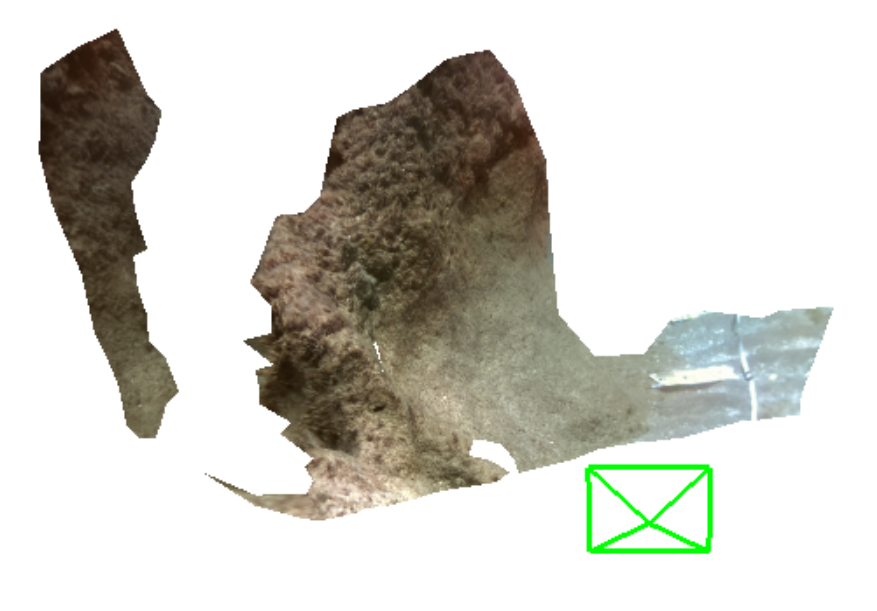}
        \label{fig:outlier:after}
    }
    \caption{Comparison of 3D mesh maps. (a) Before outlier rejection. (b) After outlier rejection.}
    \label{fig:outlier}
\end{figure}

For real-time mapping, the proposed method first performs point sampling and then performs Delaunay triangulation \cite{dt} to generate 2D mesh information using the points as vertices of a triangle. In this paper, we extract the feature points \cite{gftt} of the image as a point sampling method so that the generated 2D mesh information can reflect the planar information of the real environment as much as possible. In addition, Delaunay triangulation is a method that generates triangles that are as close to equilateral triangles as possible so the 2D mesh information can maintain the flatness and texture of the image as much as possible, even when projected to 3D. The results of the 2D mesh generation can be seen in \figref{fig:triangulation}.
The vertices of the 2D meshes are then projected into the 3D using the depth information, and a 3D mesh map is generated. The result can be seen in \figref{fig:outlier:before}.

\subsubsection{Outlier Rejection}

The generation of 2D meshes can result in meshes that do not reflect the real 3D planarity. To solve this problem, we perform outlier rejection in terms of 2D and 3D meshes. First, after generating a 2D mesh, we remove triangles whose length of one side of the triangle is longer than a certain value according to the condition in \equref{equ:2dod}. 

\begin{equation}\label{equ:2dod}
    \max(
        \Vert {\mathbf{p}_2 - \mathbf{p}_1} \Vert,
        \Vert \mathbf{p}_3 - \mathbf{p}_2 \Vert, 
        \Vert \mathbf{p}_1 - \mathbf{p}_3 \Vert, 
    )
    >
    l_p
\end{equation}
where $\mathbf{p}_1, \mathbf{p}_2,$ and $\mathbf{p}_3$ are three vertices of a triangle in 2D image plane and $l_p$ is the threshold length of one side of the triangle. The red-colored triangles in \figref{fig:triangulation} are the ones filtered by the above equation.

Next, we project the meshes into the 3D space and perform outlier rejection. In a similar scheme to \equref{equ:2dod}, we exclude triangles with sides of large length. This is done by the following \equref{equ:3dod1}.
\begin{equation}\label{equ:3dod1}
    \max(
        \Vert {\mathbf{v}_2 - \mathbf{v}_1} \Vert,
        \Vert \mathbf{v}_3 - \mathbf{v}_2 \Vert, 
        \Vert \mathbf{v}_1 - \mathbf{v}_3 \Vert, 
    )
    >
    l_v
\end{equation}
where $\mathbf{v}_1, \mathbf{v}_2,$ and $\mathbf{v}_3$ are three vertices of a triangle in 3D space and $l_v$ is the threshold length of one side of the triangle. 

Then, if the angle between the camera to the center of the mesh and the normal vector of the mesh is close to vertical, we remove them using \equref{equ:3dod2}. 
\begin{equation}\label{equ:3dod2}
    \vert \mathbf{v}_c \cdot \mathbf{n} \vert < d
\end{equation}
The mesh that is removed using \equref{equ:3dod1} and \equref{equ:3dod2} is the purple triangles in \figref{fig:triangulation}. And the 3D mesh map before and after removing outliers can be seen in \figref{fig:outlier}.

\subsubsection{Map Expansion}

\begin{figure}[t] 
    \centering
    \includegraphics[width=\columnwidth]{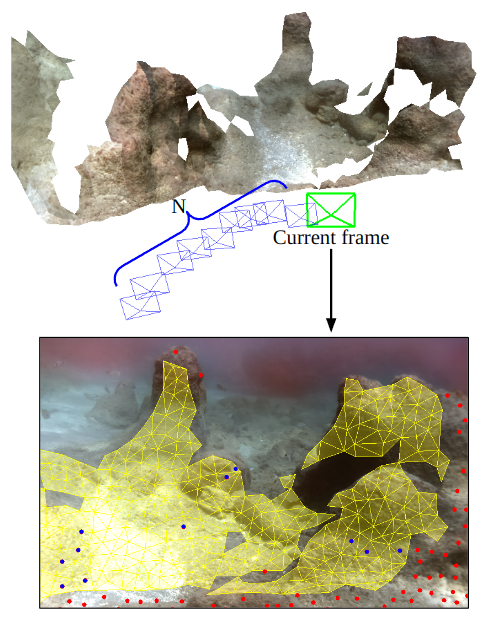}
    \caption{Mesh map expansion: (up) 3D mesh map generated from the previous N camera frame. (down) Yellow triangles are projected mesh from a global map. Red dots are sampled points found in the current frame that are outside the mesh. Blue points are sampled points that are above the existing mesh in the image plane but have a significant distance to the mesh in 3D.}
    \label{fig:expansion}
\end{figure}

To expand the map in real-time, the proposed method employs a sliding window strategy that first projects the 3D mesh map generated in the previous N frames onto the current image plane and then expands the map. At this time, we sample points that are farther away from the projected vertices than the minimum distance. The obtained points fall into one of two cases: (1) Inside the existing mesh (2) Outside the existing mesh. In the case of (2), the information of the sampled points is saved for map expansion. In the case of (1), the distance between the point and the plane is calculated as shown in \equref{equ:ppdist} and stored only if the distance is greater than a certain value.
\begin{equation}\label{equ:ppdist}
    \frac{(\mathbf{v}_2-\mathbf{v}_1)\times(\mathbf{v}_3-\mathbf{v}_1)}
{\vert(\mathbf{v}_2-\mathbf{v}_1)\times(\mathbf{v}_3-\mathbf{v}_1)\vert}
    \cdot(\mathbf{v}_n-\mathbf{v}_1) < d_{\min}
\end{equation}
where $\mathbf{v}_n$ is the 3D position of the sampled point and $d_{\min}$ is the minimum distance between the new point and the mesh plane. Finally, the map is expanded by Delaunay triangulation again, including the existing vertices and the newly obtained points. \figref{fig:expansion} shows the map expansion process.

\begin{figure*}[t!] 
    \centering
    \includegraphics[width=0.9\textwidth]{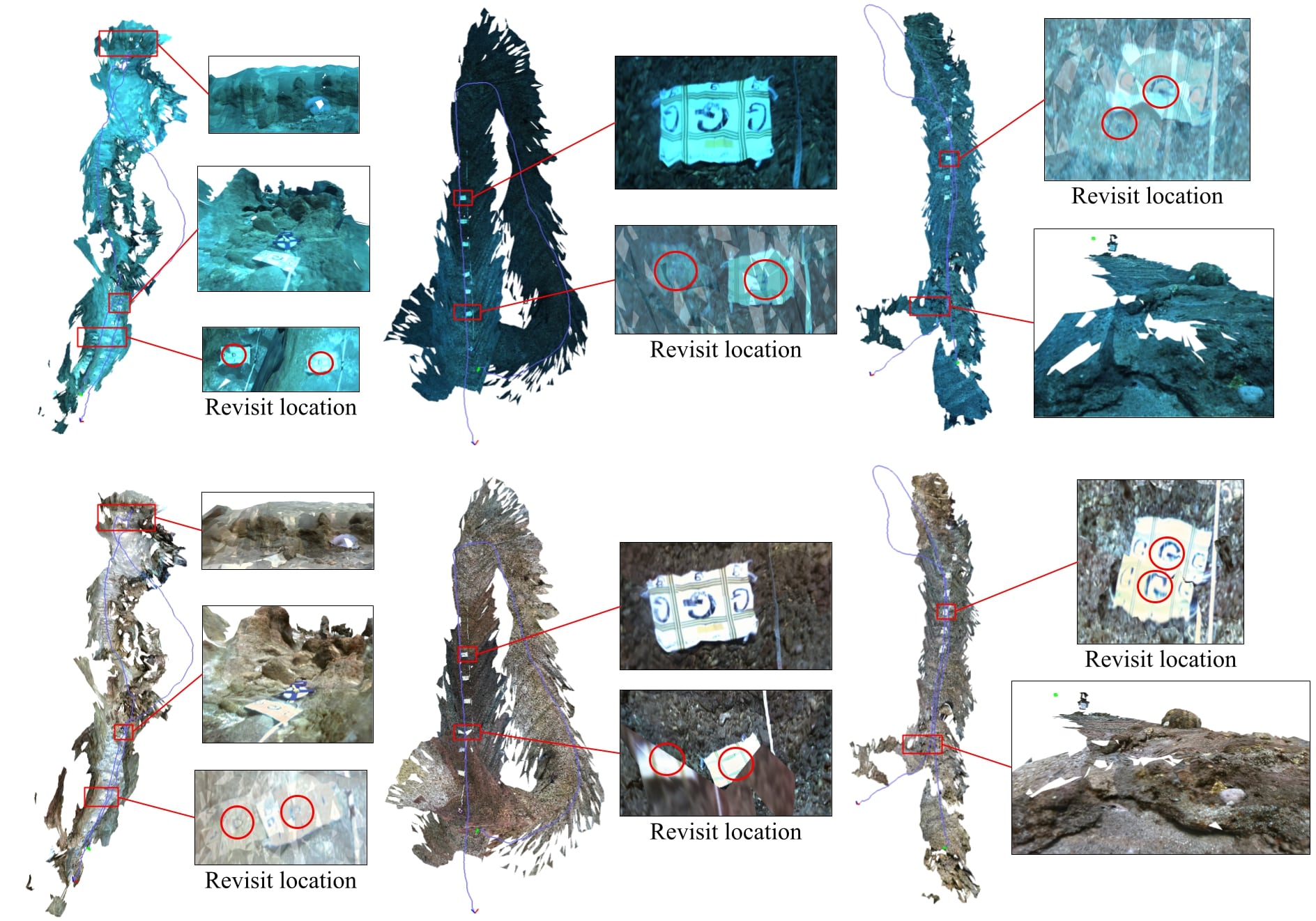}
    \caption{Mapping result with/without image enhancement: (up) without image enhancement, (down) with image enhancement, (left) \textit{FLSea-VI Canyons Flatiron}, (middle) \textit{FLSea-VI Redseas Cross Pyramid Loop}, (right) \textit{FLSea-VI Redseas Northeast Path}.}
    \label{fig:result_main}
\end{figure*}

\section{Experimental Results}

\subsection{Setting}
This paper utilizes the FLSea dataset \cite{flsea} for qualitative validation of the proposed method. It provides data acquired with stereo vision and visual-inertial sensors. While most underwater datasets provide bottom-facing images, this dataset provides forward-facing images. The dataset also provides 3D information in the form of depth images generated by the \ac{SfM} method using the commercial program Agisoft Metashape. In this paper, we verify the proposed method using the images and depth information provided by the dataset. Furthermore, to compare the quantitative results of the proposed method, the image from the indoor synthetic dataset ICL-NUIM \cite{iclnuim} was modeled as an underwater image. The experiments were conducted on an Intel Core i5-13400F CPU and run on Ubuntu 20.04.

\begin{figure}[t] 
    \centering
    \includegraphics[width=\columnwidth]{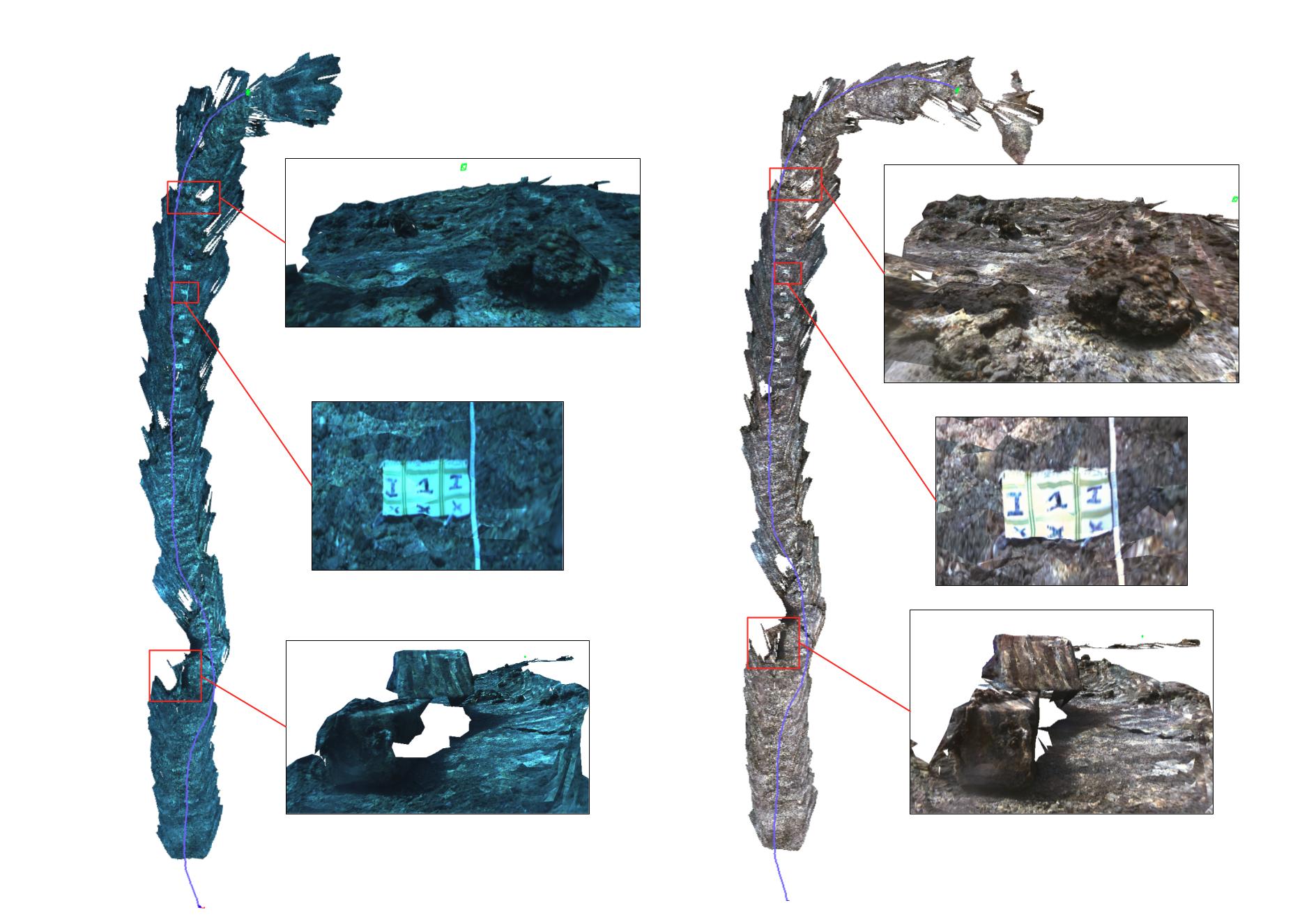}
    \caption{Mapping result for \textit{FLSea-VI Redseas Big Dice Loop} sequence without image enhancement (left) and with image enhancement (right).}
    \label{fig:result_main_added}
\end{figure}

\subsection{Real-world Dataset}

\subsubsection{Localization}

Due to the characteristics of the underwater environment, it is difficult to obtain ground truth trajectories. Since the datasets we utilized also do not provide ground truth trajectories, we conducted a qualitative verification of the improvement in localization with image enhancement.

The left sequence \textit{FLSea-VI Canyons Flatiron} and the right sequence \textit{FLSea-VI Redseas Northeast Path} in \figref{fig:result_main} have a reverse loop closure. The red circles in “Revisit location” in \figref{fig:result_main} represent the same landmarks, and the maps are based on images acquired looking in opposite directions. This result shows the distance between the same landmarks is closer on the created map than without image enhancement. It means that the drift error, which is caused by the accumulation of errors in the process, is reduced after image enhancement.

By checking the revisited location of the \textit{FLSea-VI Redseas Cross Pyramid Loop} in the center of \figref{fig:result_main}, the locations of the artificial markers are not significantly different. This is because image enhancement affects the extraction and comparison of feature points in the localization process, but the same depth information is used regardless of image enhancement. If image-based depth estimation is performed, it can be expected that the improvement in localization will be more emphasized with image enhancement.

In the \textit{FLSea-VI Redseas Big Dice Loop data} in \figref{fig:result_main_added}, the robot fails to localization at some point. The map generated until it failed localization can be seen in In \figref{fig:result_main_added}. The robot's localization failure is expected to be due to its fast rotational motion at that particular point in the sequence, which caused it to fail to track the feature. The results on the right side of \figref{fig:result_main_added} show that the robot could track farther than without image enhancement. This suggests that image enhancement helps the robot to perform localization robustly.

\begin{figure}[t] 
    \centering
    \includegraphics[width=\columnwidth]{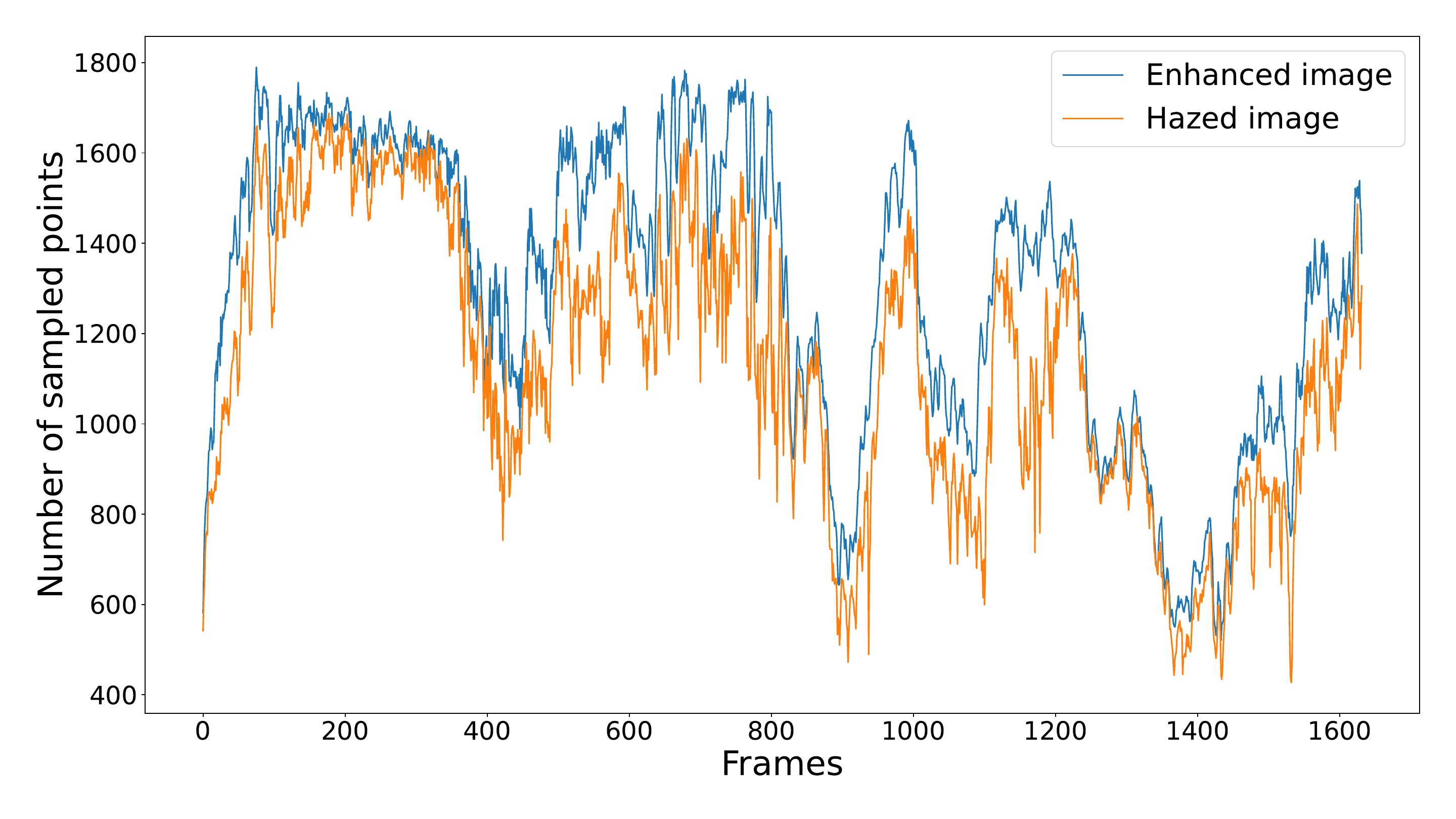}
    \caption{Number of sampled points for hazed and enhanced image.}
    \label{fig:sampling_result}
\end{figure}

\begin{figure}[t] 
    \centering
    \subfigure[Mesh.]{
        \includegraphics[width=0.28\columnwidth]{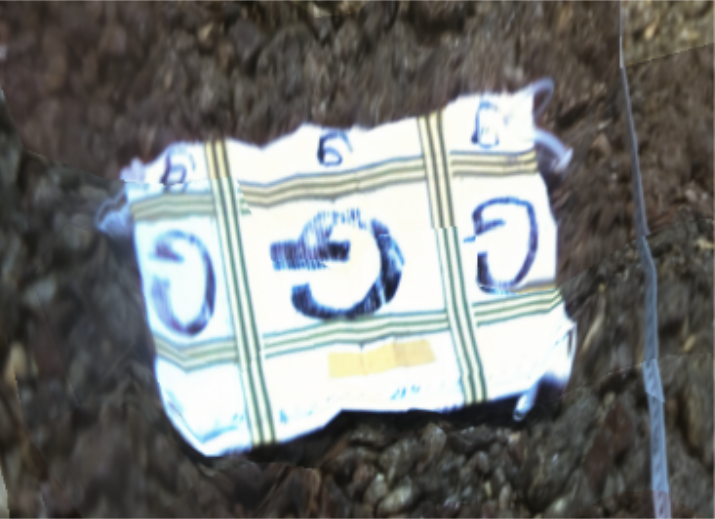}
        \label{fig:maprep:mesh}
    }
    \subfigure[Point cloud.]{
        \includegraphics[width=0.28\columnwidth]{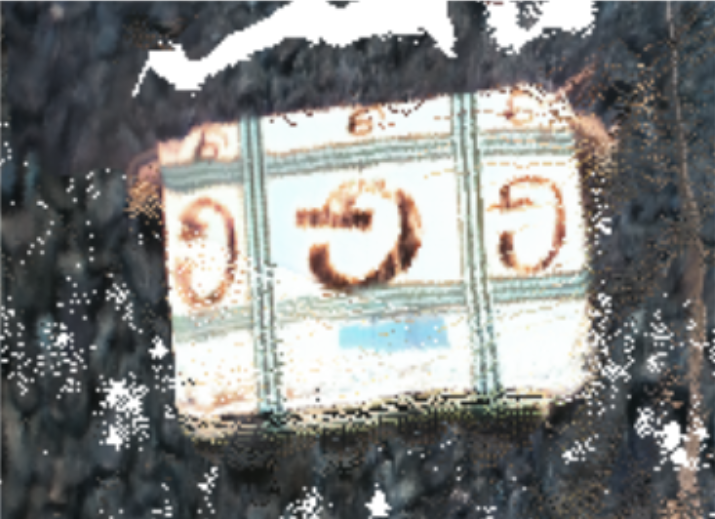}
        \label{fig:maprep:pcl}
    }
    \subfigure[Voxel.]{
        \includegraphics[width=0.28\columnwidth]{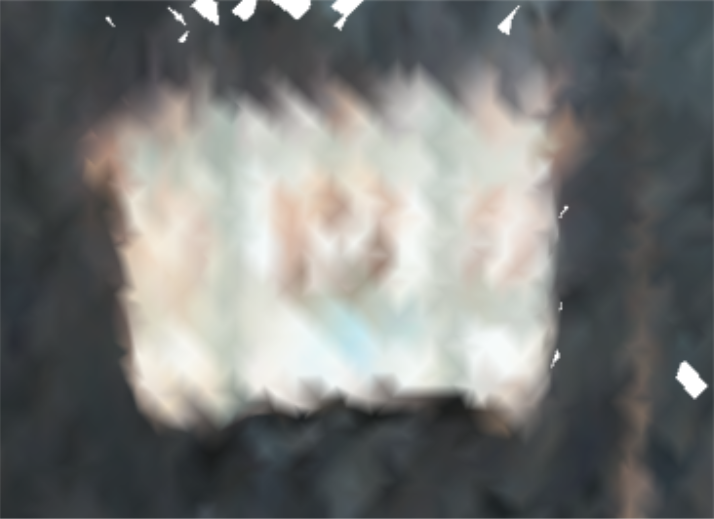}
        \label{fig:maprep:voxel}
    }
    \caption{Comparison of qualitative results using artificial landmarks in the \textit{FLSea-VI Redseas Cross Pyramid Loop} sequence. (a) Proposed method. (b) Point cloud. (c) Voxblox.}
    \label{fig:compmaprep}
\end{figure}

\subsubsection{Point Sampling}

In this paper, we compare the performance of point sampling with and without image enhancement in \figref{fig:sampling_result}. Since the proposed method performs point sampling based on the feature point extraction \cite{gftt}, it will show better performance after improving the color distortion and contrast of the image. To compare the performance of point sampling, the number of feature points detected in the image of \textit{FLSea-VI Redseas Cross Pyramid Loop} was verified before and after image enhancement. We found that the number of feature points detected after the image enhancement increased by about 16.1\% compared to before the enhancement.

\subsubsection{Photorealistic Mapping}

The proposed method enables realistic mapping by texturing the mesh. This is important for human interpretation of the mapping results after \ac{AUV} performs their tasks. The realistic mapping by image enhancement can be seen in \figref{fig:result_main}, \figref{fig:result_main_added}, \figref{fig:result_extra1} and \figref{fig:result_extra2}.

To check the interpretability of the maps, we verified whether humans can recognize the writing of artificial markers present in the \textit{FLSea-VI Redseas Cross Pyramid Loop}. We compared the map rendered using Mesh with maps generated from point clouds and voxels, commonly used in existing methodologies. Point cloud maps were generated directly using depth and color information from keyframes in ORB-SLAM2 \cite{orbslam2}, and voxel-based maps were generated by applying the marching cube algorithm to TSDF maps obtained at 1(cm) using the Voxblox \cite{voxblox} algorithm. 

As in \figref{fig:compmaprep}, mesh and point cloud results are enough to recognize the text. In the case of voxels, it was not sufficiently detailed to recognize the text, even though voxels were as small as 1(cm). Despite the good results of the point cloud, it requires a large data size to represent a map at a level that humans can interpret because one point represents only one color. It can be shown in \tabref{tab:compmappoint}.

\begin{table}
\caption{Number of point(voxel) in \textit{FLSea-VI Redseas Cross Pyramid Loop} sequence.}
\def\arraystretch{1.5}
\begin{center}
\begin{tabular}{ L{17mm} | C{17mm} | C{17mm} | C{17mm} }
    \hline
        & Mesh & Point cloud & Voxel (1cm) \\
    \hline\hline
    Number of point(voxel)   & 17,538        & 22,833,823        & 13,997 \\
    \hline
\end{tabular}
\end{center}
\label{tab:compmappoint}
\end{table}

\begin{figure}[t] 
    \centering
    \includegraphics[width=\columnwidth]{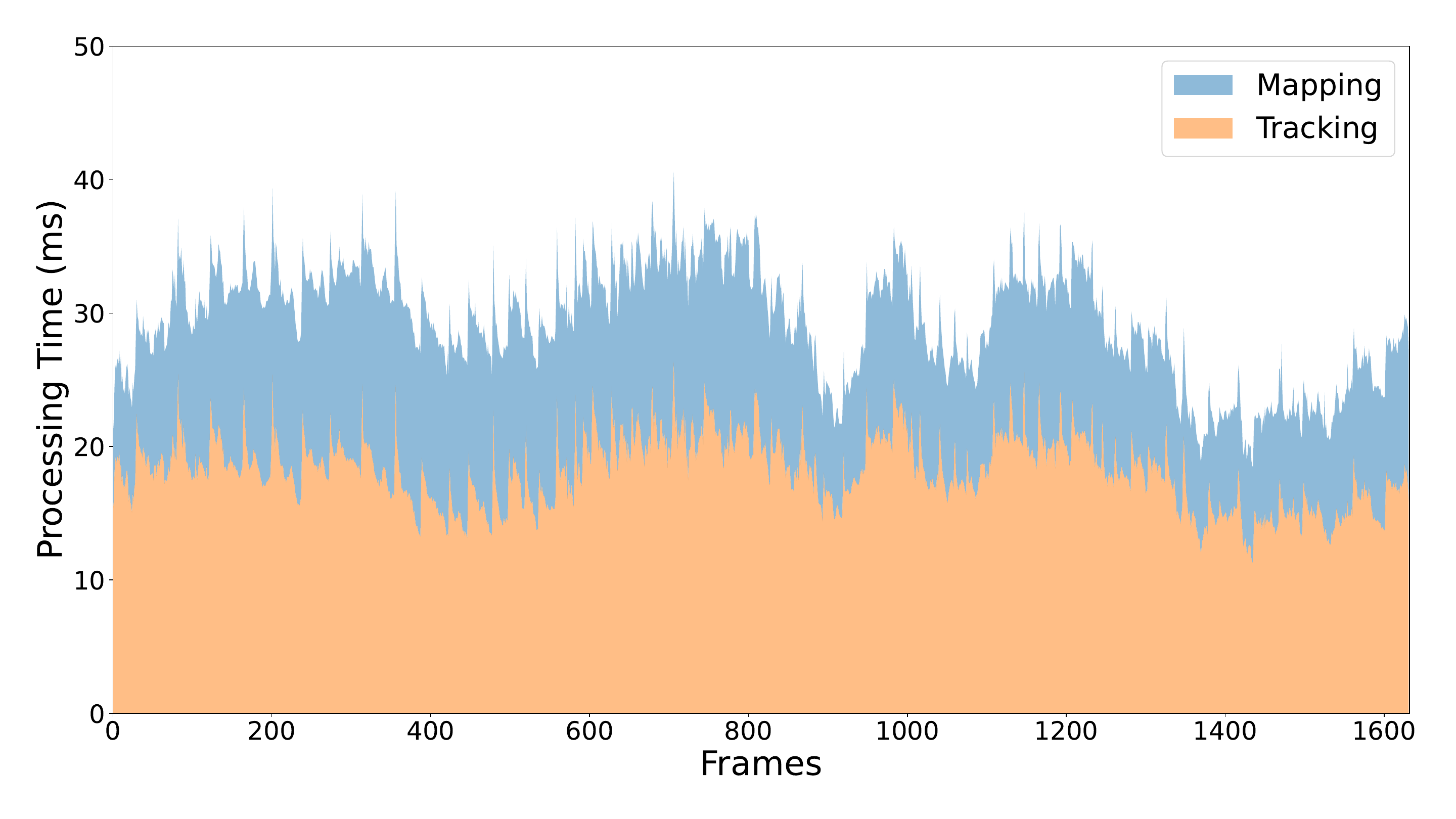}
    \caption{Processing time for \textit{FLSea-VI Redseas Cross Pyramid Loop} sequence: The orange area represents tracking time, and the blue area represents mapping time.}
    \label{fig:time}
\end{figure}

\subsubsection{Data Size and Processing Time}

\begin{figure*}[t!] 
    \centering
    \includegraphics[width=\textwidth]{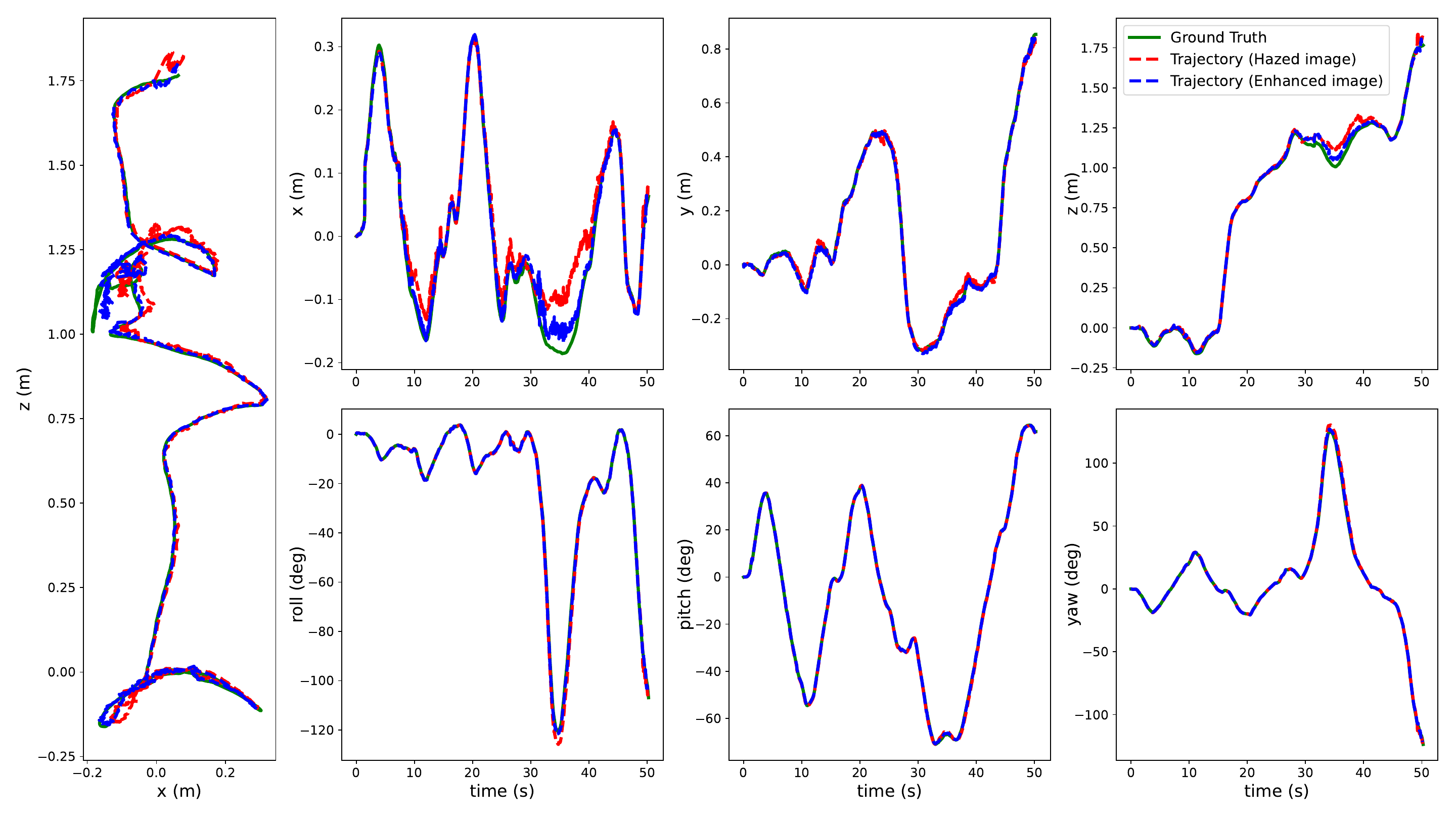}
    \caption{Trajectory, position, and orientation of the results compared to the ground-truth using the hazed and enhanced image for \textit{of kt0} sequence.}
    \label{fig:quan_result}
\end{figure*}

\begin{figure}[t] 
    \centering
    \subfigure[Hazed image.]{
        \includegraphics[width=0.29\columnwidth]{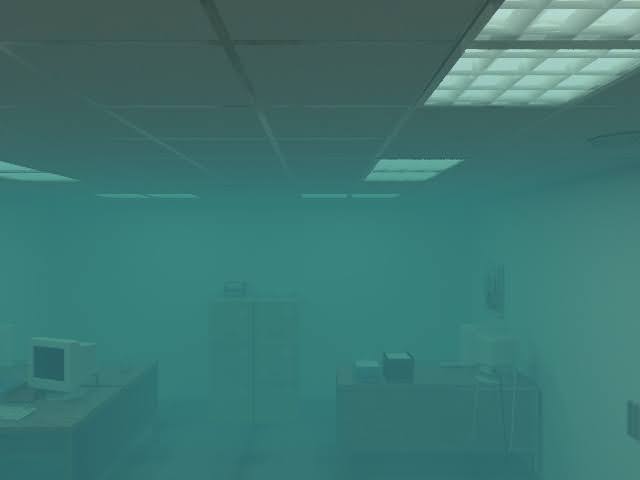}
        \label{fig:synth:hazed}
    }
    \subfigure[Enhanced image.]{
        \includegraphics[width=0.29\columnwidth]{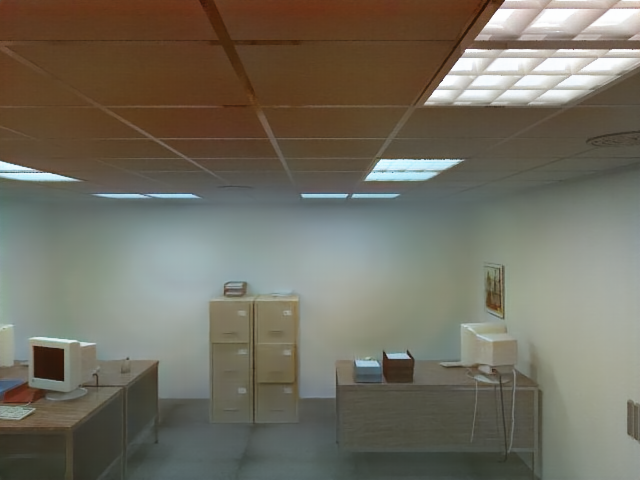}
        \label{fig:synth:enhanced}
    }
    \subfigure[Original image.]{
        \includegraphics[width=0.29\columnwidth]{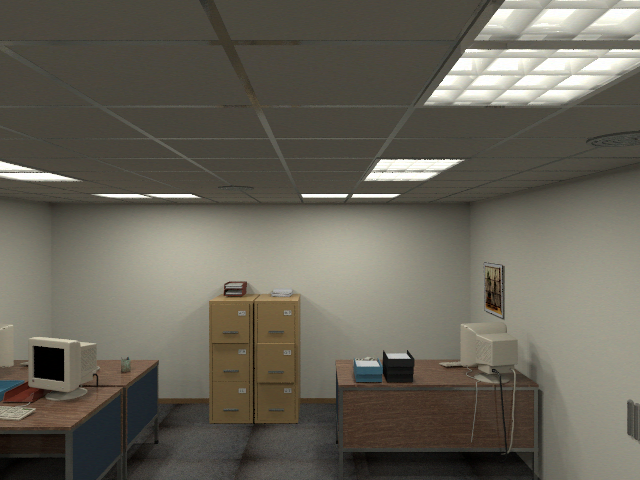}
        \label{fig:synth:original}
    }
    \caption{ The original image of the dataset (c), the distorted image following the underwater light propagation model\cite{mclpm} (a), and the enhanced image using Joint-ID \cite{jointid}.}
    \label{fig:synth_uw_model}
\end{figure}

In \tabref{tab:compmappoint}, it shows how many points or voxels are stored according to the map representation type. When using the mesh representation, it stores significantly fewer points than using point cloud while still showing a high-quality map as shown in \figref{fig:compmaprep}. When voxel-based representation is used, it is possible to represent the map with fewer voxel blocks than the mesh-based map, but it is not possible to create a realistic map that is sufficient for human interpretation. 

The proposed method achieves real-time mapping by using mesh-based representation and a sliding window map expansion strategy. In this paper, we evaluated the proposed method by setting $N=25$ as the window size and using images with $720\times456$ resolution. The results in \figref{fig:time} illustrate the processing time per frame. The results demonstrate that the ORB-SLAM2-based localization took an average of 18.14(ms), and the mapping took an average of about 11.18(ms). The overall processing time per frame is about 29.32(ms) on average, which shows that the proposed method can operate in real time in terms of localization and mapping.

\subsection{Synthetic Dataset}

Due to the characteristics of the underwater environment, it is difficult to obtain ground truth trajectory, and also the dataset used in this paper does not provide ground truth trajectory information. Therefore, to perform a quantitative evaluation of the proposed method, we simulate the underwater environment by distorting the image data of ICL-NUIM \cite{iclnuim}, an indoor synthetic dataset. We use the commonly used McGlamery's underwater light propagation model \cite{mclpm} to represent the distorted underwater image from the original image. The original and the distorted underwater image can shown in \figref{fig:synth_uw_model}.

\begin{table}
\caption{\ac{ATE} (m) for \textit{of kt0} sequence.}
\def\arraystretch{1.5}
\begin{center}
\begin{tabular}{ L{22mm} | C{15mm} | C{15mm} | C{15mm} }
    \hline
        & Mean & Median & RMSE \\
    \hline\hline
    without image enhancement   & 0.038324 &  0.020187 & 0.055224 \\
    \hline
    with image enhancement   & \textbf{0.018117} & \textbf{0.012457} & \textbf{0.024910} \\
    \hline
\end{tabular}
\end{center}
\label{tab:ATE}
\end{table}

\figref{fig:quan_result} and \tabref{tab:ATE} show the localization results before and after image enhancement for \textit{of kt0} sequence. In this paper, we used EVO \cite{grupp2017evo} to calculate the \ac{ATE}. After image enhancement, the overall localization results are better, as shown in \figref{fig:quan_result}. There is a part of the \textit{of kt0} sequence where the camera is looking at the ceiling, and due to the lack of texture in the image, the position error is large across the results. \tabref{tab:ATE} shows that the RMSE of \ac{ATE} before and after image enhancement is 0.055224(m) and 0.024910(m), respectively, which is about 2.22 times smaller after image enhancement.

\begin{figure}[t!] 
    \centering
    \includegraphics[width=\columnwidth]{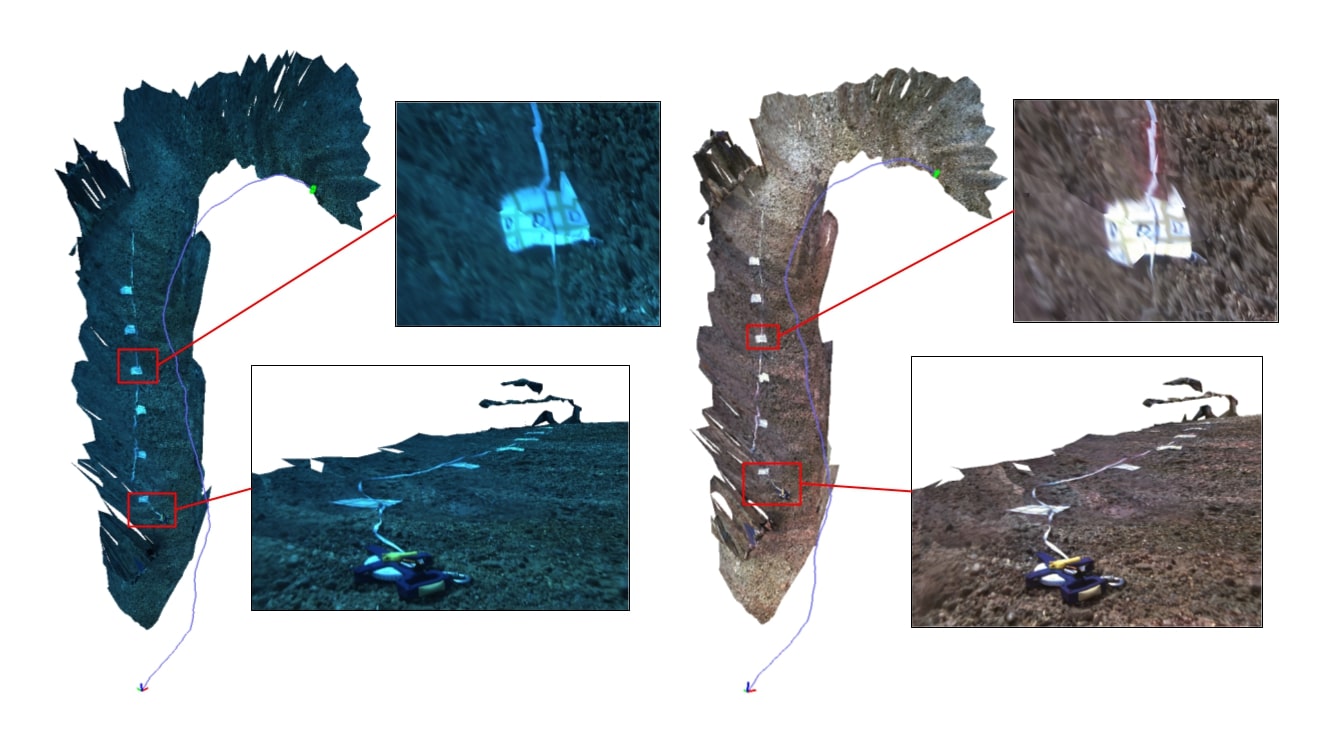}
    \caption{Mapping result for \textit{FLSea-VI Redseas Coral Table Loop} sequence without image enhancement (left) and with image enhancement (right).}
    \label{fig:result_extra1}
\end{figure}

\begin{figure}[t!] 
    \centering
    \includegraphics[width=\columnwidth]{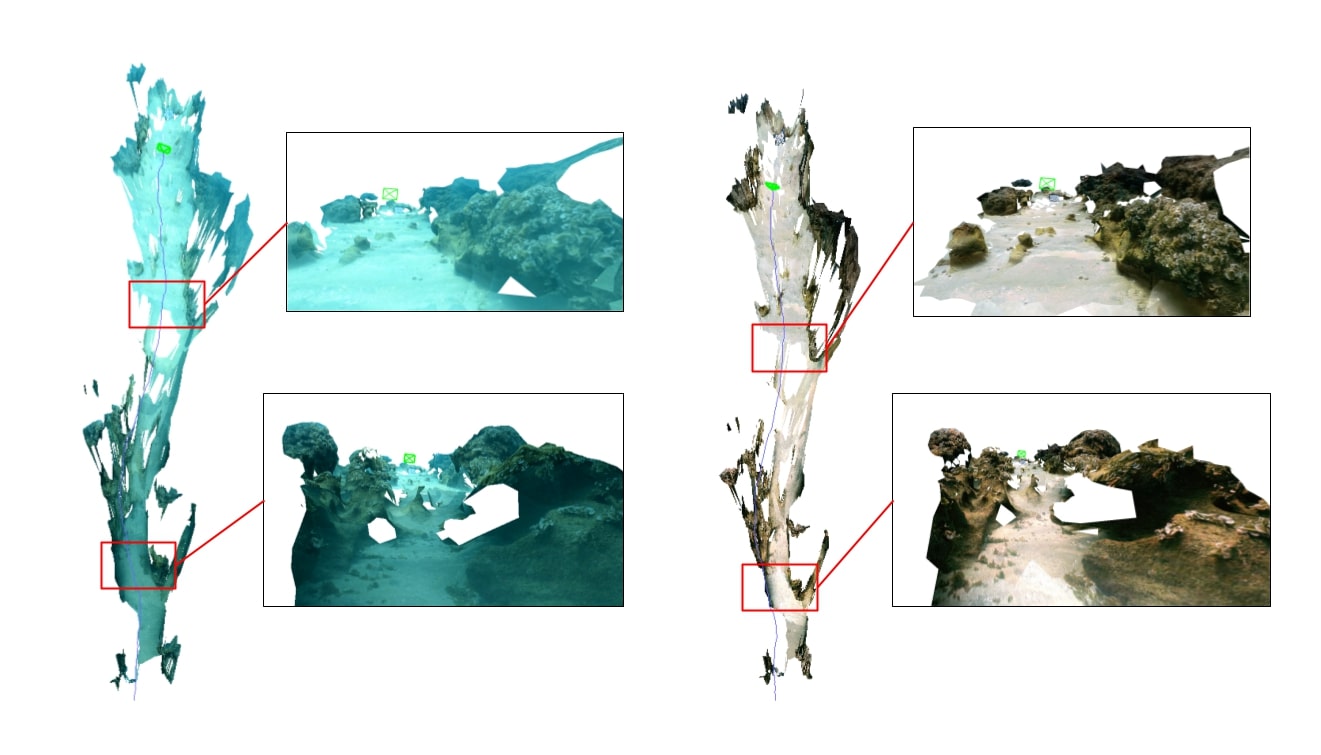}
    \caption{Mapping result for \textit{FLSea-Stereo Rock Garden 1} sequence without image enhancement (left) and with image enhancement (right).}
    \label{fig:result_extra2}
\end{figure}

\section{Conclusion}

In this paper, we propose a system for improving localization accuracy and photorealistic maps in real-time using image enhancement and mesh representation. We qualitatively verified that the robot's localization performance and map representation are improved when using the enhanced image compared to hazed image, and the created maps are easy to understand by humans. We also present additional results for qualitative verification of map creation in \figref{fig:result_extra1} and \figref{fig:result_extra2}. To quantitatively validate the effect of image enhancement, we use an augmented underwater image from the indoor synthetic dataset. It clearly shows that image enhancement improves localization performance. In the future, we will improve the performance of localization using the created mesh map data and design a system that can perform localization correction through loop closure detection.

%


\bibliographystyle{packages/IEEEtranN} 
\bibliography{packages/string-short, packages/references}

\end{document}